\documentclass[lettersize,journal]{IEEEtran}

\usepackage[utf8]{inputenc} 
\usepackage[T1]{fontenc}    
\usepackage{hyperref}
\usepackage{url}  
\usepackage{booktabs}
\usepackage{amsfonts}  
\usepackage{nicefrac} 
\usepackage{microtype}   
\usepackage{fancyhdr}      
\usepackage{bm}
\usepackage{bbm}
\usepackage{amsmath}
\usepackage{algorithmic}
\usepackage{algorithm}
\usepackage{multirow}
\usepackage{graphicx}
\usepackage{subfigure}
\usepackage{wrapfig}
\usepackage{amsthm}
\usepackage{multicol}
\usepackage{amssymb}
\usepackage{pdfpages}
\usepackage{xspace}
\usepackage{float}
\usepackage{threeparttable}
\usepackage{soul}
\usepackage{cite}
\usepackage{colortbl}
\usepackage{epsfig}
\usepackage{amssymb}
\usepackage{multirow}
\usepackage{tabularx}
\usepackage{pifont}
\usepackage{stfloats}

\makeatletter
\DeclareRobustCommand\onedot{\futurelet\@let@token\@onedot}
\def\@onedot{\ifx\@let@token.\else.\null\fi\xspace}

\def\eg{\emph{e.g}\onedot} 
\def\ie{\emph{i.e}\onedot}

\def\etal{\emph{et al}\onedot}
\makeatother

\begin{document}

\title{Improving Audio-Visual Video Parsing with \\Pseudo Visual Labels}

\author{Jinxing Zhou,
        Dan Guo,
        Yiran Zhong,
        and~Meng Wang,~\IEEEmembership{Fellow,~IEEE}

\thanks{
Jinxing Zhou is with the Key Laboratory of Knowledge Engineering with Big Data (HFUT), Ministry of Education, School of Computer Science and Information Engineering (School of Artiﬁcial Intelligence), Hefei University of Technology (HFUT), and Intelligent Interconnected Systems Laboratory of Anhui Province (HFUT), Hefei, 230601, China. 
E-mail: zhoujxhfut@gmail.com.}
\thanks{
Dan Guo and Meng Wang are with the Key Laboratory of Knowledge Engineering with Big Data (HFUT), Ministry of Education, School of Computer Science and Information Engineering (School of Artiﬁcial Intelligence), Hefei University of Technology (HFUT), Intelligent Interconnected Systems Laboratory of Anhui Province (HFUT), and Hefei Comprehensive National Science Center, Hefei, 230601, China.
E-mail: guodan@hfut.edu.cn, eric.mengwang@gmail.com.}
\thanks{Yiran Zhong is with Shanghai AI Lab, Shanghai, 200000, China.
E-mail: zhongyiran@gmail.com.}
\thanks{Corresponding authors: Dan Guo and Meng Wang.}
}

\markboth{Journal of \LaTeX\ Class Files,~Vol.~14, No.~8, August~2021}%
{Shell \MakeLowercase{\textit{et al.}}: A Sample Article Using IEEEtran.cls for IEEE Journals}


\maketitle

\begin{abstract}
Audio-Visual Video Parsing is a task to predict the events that occur in video segments for each modality. It often performs in a weakly supervised manner, where only video event labels are provided, \ie, the modalities and the timestamps of the labels are unknown. Due to the lack of densely annotated labels, recent work attempts to leverage pseudo labels to enrich the supervision. A commonly used strategy is to generate pseudo labels by categorizing the known event labels for each modality. However, the labels are still limited to the video level, and the temporal boundaries of event timestamps remain unlabeled. In this paper, we propose a new pseudo label generation strategy that can explicitly assign labels to each video segment by utilizing prior knowledge learned from the open world. Specifically, we exploit the CLIP model to estimate the events in each video segment based on visual modality to generate segment-level pseudo labels. A new loss function is proposed to regularize these labels by taking into account their category-richness and segment-richness. A label denoising strategy is adopted to improve the pseudo labels by flipping them whenever high forward binary cross entropy loss occurs. We perform extensive experiments on the LLP dataset and demonstrate that our method can generate high-quality segment-level pseudo labels with the help of our newly proposed loss and the label denoising strategy. Our method achieves state-of-the-art audio-visual video parsing performance.
\end{abstract}


\section{Introduction}\label{sec:introduction}
\IEEEPARstart{A}{coustic} and visual signals flood our lives in abundance, and each signal may carry various events.
For example, we often see cars and pedestrians on the street. Meanwhile, we can also hear the beeping of the car horns and the sound of people talking.
Humans achieve such comprehensive scene understanding in large part thanks to the simultaneous use of their auditory and visual sensors.
To imitate this kind of intelligence for machines, many research works started from some fundamental tasks of single modality understanding, such as the audio event classification~\cite{hershey2017vggish,kong2018audio,kumar2018knowledge,gong2021ast}, video classification~\cite{karpathy2014large,long2018attention,long2018multimodal,tran2019video}, and temporal action localization~\cite{zeng2019graph,chao2018rethinking,zhu2021enriching,gao2022fine}.
The audio event
classification task focuses on the recognition of the audio modality, while the video classification and action localization tasks focus on the visual modality.
With the deepening of research, many works have further explored the multi-modal audio-visual perception~\cite{wei2022learning}, giving birth to tasks such as sound source localization~\cite{arandjelovic2017look,arandjelovic2018objects,rouditchenko2019self,senocak2018learning,hu2020discriminative,hu2019deep,qian2020multiple,zhao2018sound,afouras2020self,zhou2022avs,zhou2023avss}, audio-visual event localization~\cite{tian2018audio,lin2019dual,wu2019dual,xu2020MM,xuan2020cross,zhou2021psp,mahmud2022ave,rao2022dual,xia2022cross,wu2022span,zhou2022cpsp,wang2022semantic}, and audio-visual question answering~\cite{yun2021pano,li2022learning,yang2022avqa}.

\begin{figure}[t]
  \centering
  \includegraphics[width=0.48\textwidth]{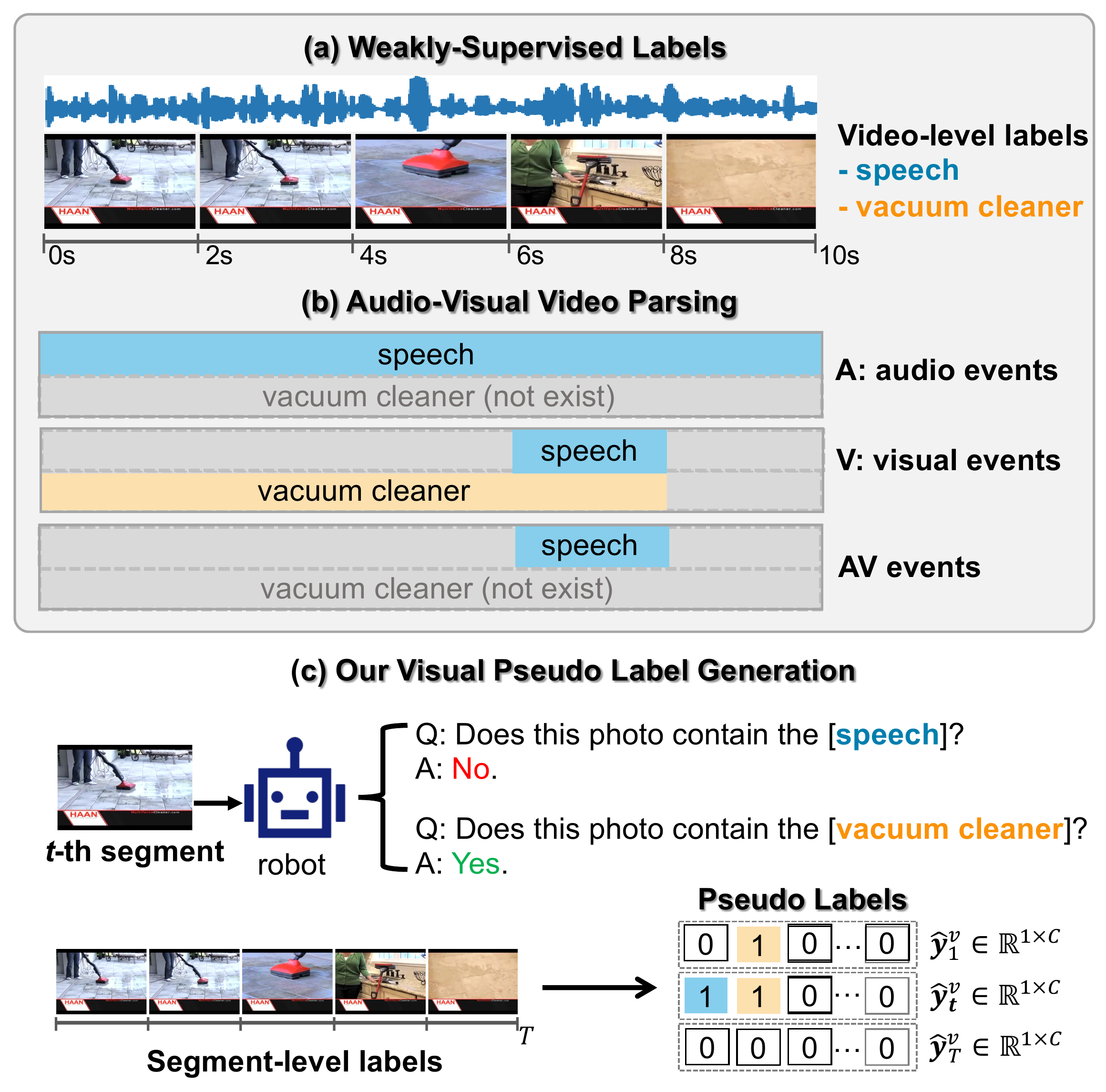}
  \caption{\textbf{An illustration of the weakly supervised Audio-Visual Video Parsing (AVVP) task and our pseudo label generation process.} 
  (a) Given a video and its event label (\textit{``speech"} and \textit{``vacuum cleaner"}),  (b) AVVP needs to predict the audio events, visual events, and audio-visual events.
  Note that \textit{"vacuum cleaner"} only exists in the visual track, while \textit{"speech"} exists in both audio and visual tracks, resulting in the audio-visual event \textit{"speech"}.
  (c) In our visual pseudo label generation process, the CLIP is used to tell what events occur in each video segment.
  We then obtain reliable segment-level visual pseudo labels to ease the weakly-supervised   video parsing task.}
  \label{fig:task_illustration}
\end{figure}

Recently, Tian \etal~\cite{tian2020HAN} proposed a new multi-modal scene understanding task, namely Audio-Visual Video Parsing (AVVP).
Given an audible video, the AVVP task asks to identify what events occur in the audio and visual tracks and in which video segments these events occur.
Accordingly, the category and temporal boundary of each event are expected to predict for each modality.
Note that both the audio and visual tracks may contain multiple distinct events and the event usually exists in consecutive segments, it is labor-intensive to provide segment-level event labels for each modality.
Consequently, AVVP is performed in a weakly supervised setting where only the video label is provided.
As the example shown in Fig.~\ref{fig:task_illustration} (a), we only know that this video contains the events of \textit{speech} and \textit{vacuum cleaner}.
For each event, the model needs to judge whether it exists in the audio modality (audio event), visual modality (visual event), or both (audio-visual event), and locate the specific temporal segments, as illustrated in Fig.~\ref{fig:task_illustration} (b).
Notably, audio-visual events are the intersection of audio and visual events.

In this work, we emphasize there are two main challenges in the AVVP task.
\textbf{1) Interference from the video label.}
{As the example shown in Fig.~\ref{fig:task_illustration} (b), the event \textit{speech} happens in both audio and visual modalities while the \textit{vacuum cleaner} only exists in the visual modality.
For data annotation, the audio and the visual track share the same supervision from the video label, \ie, \{\textit{speech, vacuum cleaner}\} together.
Thus, during the model training process, the visual object \textit{vacuum cleaner} will interfere with the audio event parsing.
Similarly, the video label may also bring interference with the visual event parsing.
} 
\textbf{2) Temporal segment distinction. }
Assuming we identify there is an event \textit{vacuum cleaner} in the visual modality, 
it is still hard to distinguish which segments contain this event (segment level) under weakly-supervised labels (video level).
These challenges make the AVVP an intractable Multi-modal Multi-Instance Learning (MMIL) problem.

In the pioneer work~\cite{tian2020HAN}, 
a Hybrid Attention Network (HAN) is proposed to encode the audio-visual temporal features and then it uses attentive pooling of the features to predict video events.
The weak video label is used as the main supervision.
To deal with the challenge, they propose to obtain the pseudo labels for separate audio and visual modalities by processing the video label with label smoothing~\cite{szegedy2016rethinking} technique. 
The experiments indicate that generating pseudo labels for each modality brings significant benefits for supervising event parsing~\cite{tian2020HAN}.
The subsequent studies diverge into two branches.
Most of them focus on \textit{designing effective networks} to implicitly aggregate the multi-modal features for prediction~\cite{mo2022MGN,pasi2022investigating,lamba2021cross,yu2022MMPyramid,lin2021CVCMS,jiang2022DHHN}, while using the pseudo labels generated by HAN~\cite{tian2020HAN}.
In contrast, the other new works~\cite{wu2021MA,cheng2022JoMOLD}
devote to \textit{generating pseudo labels} for each modality based on the baseline network of HAN.
The generated pseudo label is denoised from the known video label, hence it is limited to the video level that only indicates what events exist in each modality.
It is still difficult to distinguish which segments the event is contained in.

To deal with the above-mentioned challenges, our work starts with the intuition that can we explicitly generate pseudo labels for each segment, at least for one modality, to facilitate this MMIL task. 
This is inspired by two observations: 1) the AVVP models are expected to be well-guided with segment-level labels as the segment-level labels can provide more explicit supervision information;
2) in this multi-modal task, humans can considerably enhance video comprehension even if some small tips of one modality are provided.
Compared to the audio, the visual signal often has more explicit semantics contained in the image, making them easier to be classified even at the segment level. Moreover, in practice, it is hard to generate segment-level pseudo labels for audio modality.
The reason is that audio signal is continuous and  difficult to separate along the timeline, and the mainstream audio event classification models usually classify the category for the entire video rather than for each segment~\cite{hershey2017vggish,kong2018audio,kumar2018knowledge,gong2021ast}. 
To this end, we propose a \textbf{Visual Pseudo LAbel exploratioN (VPLAN) method} that aims to generate high-quality segment-level pseudo labels for the visual modality and further helps the multi-modal AVVP task.

To obtain the visual pseudo labels, a natural idea is to borrow free knowledge from pretrained models for image classification, such as various 
classification models or vision transformers pretrained on ImageNet~\cite{deng2009imagenet}.
Given an image extracted from 
a video sample in the AVVP task, such pretrained models can classify the image into predefined categories of the ImageNet, thereby obtaining frame-level or segment-level labels. 
However, there is a misalignment of the category vocabulary between the source and the target data.
Therefore, simply using the image classification models will introduce non-negligible label noise.
Recently, vision-language pre-training has attracted tremendous attention which learns advanced representations of both visual images and texts from large-scale web data, facilitating many cross-modal downstream tasks~\cite{alayrac2022flamingo,jia2021scaling,radford2021CLIP}.
As a representative work, the Contrastive Language-Image Pretraining (CLIP)~\cite{radford2021CLIP} opens a new paradigm for zero-shot image classification.
Given an image, 
its potential category names are inserted into a predefined text prompt.
Then CLIP can score the categories according to the similarity between the encoded texts and the image features. 
CLIP is flexible in classifying images from an open-category vocabulary.

Inspired by this, we propose a \textbf{Pseudo Label Generation (PLG)} module that utilizes the CLIP~\cite{radford2021CLIP} to generate segment-level visual pseudo labels\footnote{Unless specified otherwise, we use ``pseudo labels'' hereafter to denote the visual pseudo labels.}.
A simple illustration can be seen from Fig.~\ref{fig:task_illustration} (c).
Given all the potential event labels, CLIP acting like an intelligent robot is asked to answer whether the event is contained in the given segment.
The queried event categories with high scores are finally regarded as visual pseudo labels. 
This process can be applied to a video frame or a video segment. 
In this work, we split a video into multiple video segments and explore the segment-level labels. We provide more implementation details in Sec.~\ref{sec:method_PLG}.
The generated pseudo labels are used to provide fine-grained supervision for model training. 
In addition, we find that the obtained segment-level pseudo labels contain rich information, indicating \textit{how many categories of events happen in each segment} (category-richness) and \textit{how many segments a certain category of the event exists in} (segment-richness).
As the example shown in Fig.~\ref{fig:task_illustration} (b), the video-level label indicates that there may be at most \textit{two} events in the visual track, \ie, the \textit{speech} and \textit{vacuum cleaner}.
In practice, only the fourth segment contains both \textit{two} events while the first segment contains \textit{one} event vacuum cleaner.
Therefore, we can denote the category richness for the fourth and the first segments as 1 and 1/2, respectively.
Similarly, from the perspective of the event categories, the vacuum cleaner appears in \textit{four} video segments of the entire video (\textit{five} segments), while the speech only exists in \textit{one} (the fourth) segment.
Thus, we can denote the segment richness for events of \textit{vacuum cleaner} and \textit{speech} as 4/5 and 1/5, respectively.
The model should be aware of such differences in category richness and segment richness to give correct predictions.
Based on this, we propose a novel \textbf{Richness-aware Loss (RL)} to align the richness information contained in predictions with that contained in pseudo labels.
Our experiments verify that the generated pseudo labels combined with the RL loss boost the video parsing performance.
Lastly, considering that the pseudo labels generated by PLG are human-free and may contain some noise, we further propose a \textbf{Pseudo Label Denoising (PLD)} strategy to refine the pseudo labels.
Samples with noisy labels are usually hard to learn and get a large forward propagation loss~\cite{arpit2017closer,hu2021learning,kim2022large}.
In our work, the large loss may come from those hard-to-learn data that are assigned incorrect pseudo labels.
PLD is designed to refine the pseudo labels of these data.
We provide more implementation details in Sec.~\ref{sec:method_PLD}.

We evaluate our method on the public LLP dataset~\cite{tian2020HAN}.
The experimental results demonstrate the effectiveness of our main designs.
Our method achieves new state-of-the-art in the visual event and audio-visual event parsing while remaining competitive in audio event parsing.
To the best of our knowledge, we are the first to explore the visual pseudo labels strategy from the segment level.
It is noteworthy that the pseudo labels generated by the PLG and further refined by the PLD and the proposed RL loss can be easily applied to existing methods for the AVVP task.


\vspace{-3mm}
\section{Related Work}\label{sec:related_work}

\textbf{Audio-Visual Video Parsing (AVVP).}
AVVP task needs to recognize what events happen in each modality and localize the corresponding video segments where the events exist.
Tian \etal~\cite{tian2020HAN} first propose this task and design a hybrid attention network to aggregate the intra-modal and inter-modal features.
Also, they use the label smoothing~\cite{szegedy2016rethinking} strategy to address the modality label bias from the single video-level label.
Some methods focus on network design.
Yu \etal ~\cite{yu2022MMPyramid} propose a multimodal pyramid attentional network that consists of multiple pyramid units to encode the temporal features. 
Jiang \etal ~\cite{jiang2022DHHN} 
use two extra independent visual and audio prediction networks to alleviate the label interference between audio and visual modalities.
Mo \etal ~\cite{mo2022MGN} use learnable class-aware tokens to group the semantics from separate audio and visual modalities. 
To overcome the label interference, Wu \etal ~\cite{wu2021MA} swap the audio and visual tracks of two event-independent videos to construct new data for model training.
The pseudo labels are generated according to the predictions of the reconstructed videos. 
Cheng \etal ~\cite{cheng2022JoMOLD} first estimate the noise ratio of the video label and reverse a certain percentage of the label with large forward losses.
Although these methods bring considerable improvements, they can only generate the event label from the video level.
Unlikely, we aim to directly obtain high-quality pseudo labels for the visual modality from the segment level that further helps the whole video parsing system training.

\textbf{Learning with Pseudo Labels}.
Deep neural networks achieve remarkable performance in various tasks, largely due to the large amount of labeled data available for training.
The methods of pseudo labels aim to utilize massive and unlabeled data to further boost performance.
Generally, a knowledgeable model is first trained on the labeled data and then it is used to generate pseudo labels on unlabeled data.
The labeled and pseudo-labeled data are finally used to train a new model.
This strategy has been shown to be beneficial for various tasks, such as image classification~\cite{yalniz2019billion,xie2020self,pham2021meta,rizve2021defense,zoph2020rethinking,Hu_2021_CVPR}, speech recognition~\cite{kahn2020self,park2020improved}, and image-based text recognition~\cite{patel2023seq}.
In this work, we use a pretrained big model as the knowledgeable model to generate pseudo labels for the data used in audio-visual video parsing task, and the pseudo-labeled data are then used for better parsing model training.
The generated pseudo labels generated may be still noisy.
There are some works that explore training loss during model learning with noisy labels.
Hu \etal~\cite{hu2021learning} propose to optimize the network by giving much weight to the clean samples while less on the hard-to-learn samples.
In the weakly-supervised multi-label classification problem, Kim \etal~\cite{kim2022large} propose to correct the false negative labels that are likely to have larger losses.
Inspired by these, we design a label denoising strategy to tackle the potential noise in our generated segment-level pseudo labels.

\textbf{CLIP Pre-Training.} Here, we also discuss the pre-training technique and elaborate on the reason to choose the CLIP as the base big model for generating pseudo labels. CLIP~\cite{radford2021CLIP} is trained on a dataset with 400 million image-text pairs using the contrastive learning technique.
This large-scale pretraining enables CLIP to learn efficient representations of the images and texts and demonstrates impressive performance on zero-shot image classification.
Its zero-shot transfer ability opens a new scheme to solve many tasks and spawns a large number of research works, such as image caption~\cite{Barraco_2022_CVPR}, video caption~\cite{tang2021clip4caption}, and semantic segmentation~\cite{ma2022open,ding2022decoupling,xu2021simple,zhou2022extract,rao2022denseclip}.
Most of the works choose to freeze or fine-tune the image and text encoders of CLIP to extract advanced features for downstream tasks~\cite{tang2021clip4caption,wang2022cris,Barraco_2022_CVPR,ma2022open,zhou2022zegclip}.
For the zero-shot semantic segmentation, some methods start to use the pretrained CLIP to generate pixel-level pseudo-labels which are annotator-free and helpful~\cite{zhou2022extract,rao2022denseclip}.
In this work, we make the first attempt to borrow the prior knowledge from CLIP to ease the challenging audio-visual video parsing task.



\section{Preliminary}
\label{sec:pre}
In this section, we  formulate the detail of the AVVP task and briefly introduce the baseline framework HAN~\cite{tian2020HAN} that is used in both our approach and these pseudo label generation works~\cite{wu2021MA,cheng2022JoMOLD} in the AVVP task.

\begin{figure*}[t]
\centering
\includegraphics[width=\textwidth]{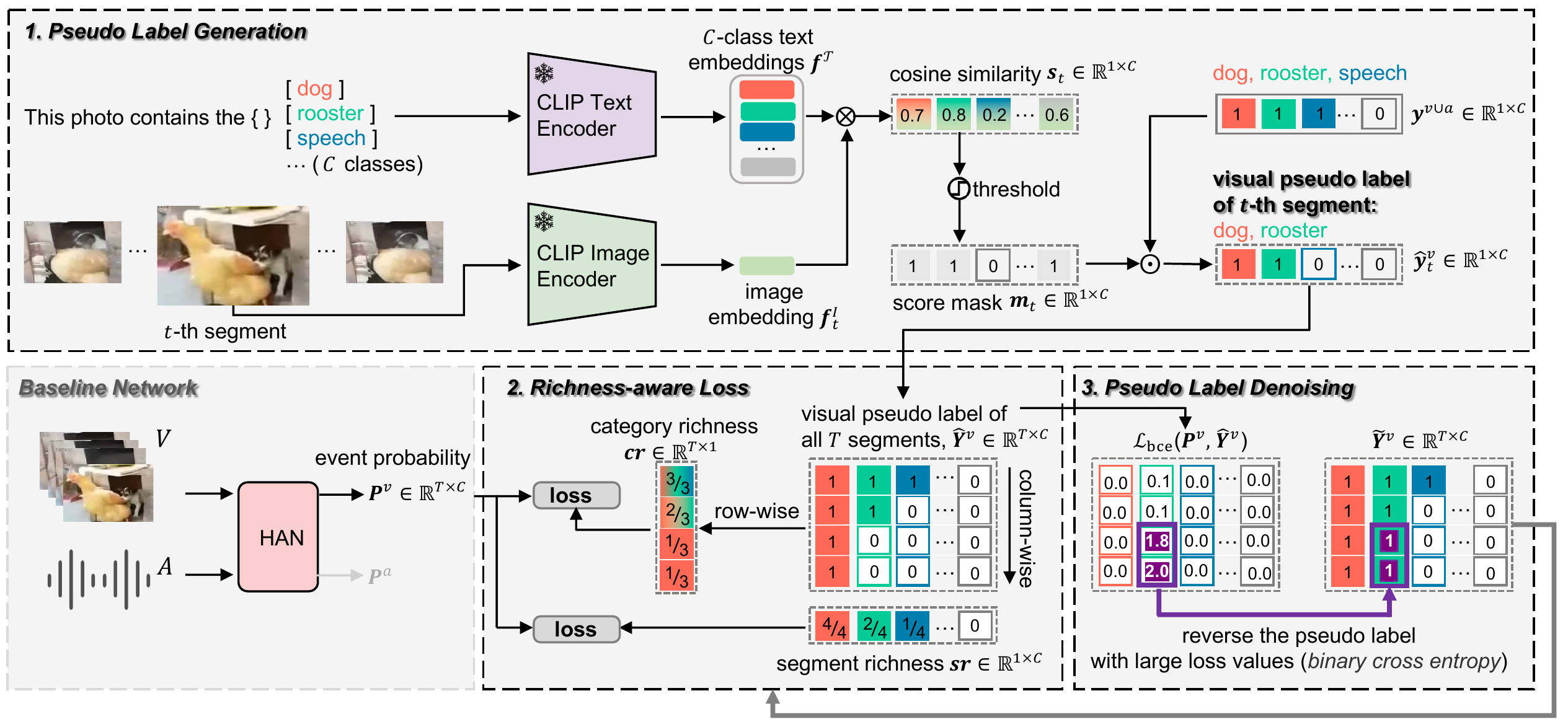}
\vspace{-3mm}
\caption{\textbf{Overview of our method.} As a label refining method, we aim to produce high-quality and fine-grained segment-wise video labels. For the backbone, any network on the AVVP task can be used to generate event predictions. Here, we adopt the baseline HAN~\cite{tian2020HAN}.
In our solution, we design a pseudo label generation module, where the pretrained CLIP~\cite{radford2021CLIP} is used to generate segment-level pseudo labels for the visual modality.
For example, for the $t$-th segment in the visual track, the video label \emph{`speech'} is filtered out for the visual modality because this category has a low similarity score calculated by CLIP. Notably, the CLIP is merely used during the pseudo label generation, and the parameters are frozen. 
After that, with the generated visual pseudo labels, we propose a richness-aware loss as a new fully-supervised supervision to help the model align the category richness and segment richness in the visual prediction and pseudo label. 
Lastly, we design a pseudo label denoising strategy that further refines the pseudo labels by reversing the positions with large forward loss values of the $\mathcal{L}_{\text{bce}}$ estimation (\ie, large binary cross-entropy). Pseudo
label refined by PLD can be further used as new supervision for the
model training.
$\otimes$ denotes the matrix multiplication and $\odot$ denotes the element-wise multiplication.}
\label{fig:framework}
\end{figure*}

\textbf{Task Formulation.} 
The AVVP task aims to locate events in video segments for audio and visual tracks and classify the event categories. 
Specifically, given a $T$-second video sequence $\{V_t, A_t\}_{t=1}^{T}$, $V_t$ and $A_t$ denote the visual and the audio components at the $t$-th video segment, respectively.
The event label of the video $ \bm{y}^{v \cup a}  \in \mathbb{R}^{1 \times C} = \{ y^{v \cup a}_c | y^{v \cup a}_c \in \{0, 1\}, c = 1, 2, ..., C\}$, 
where $C$ is the total number of event categories, the superscript $v\cup a$ denotes the event label of the entire video is the union of the labels of audio and visual modalities, value 1 of ${y}^{v \cup a}_c$ represents an event with that $c$-th category happens in the video.
Note that $\bm{y}^{v \cup a}$ is a weakly supervised label from the video level, \emph{the label of each individual modality for each video segment is unknown during training.
However, the audio events, visual events, and audio-visual events contained in each segment need to be predicted for evaluation.}
We denote the probabilities of the video-level visual and audio events as $\{\{\bm{p}^v; \bm{p}^a\} \in \mathbb{R}^{1 \times C} | {p}^v_c, {p}^a_c \in [0, 1] \}$,
$\bm{p}^{v \cap a}=\bm{p}^{v}*\bm{p}^{a}$ is used to represent the intersection of them.
Thus, the probability of the visual events, audio events, and audio-visual events of all video segments can be denoted as $\{\bm{P}^v; \bm{P}^a; \bm{P}^{v \cap a} \} \in \mathbb{R}^{T \times C} $, which need to be predicted.

\textbf{Baseline Framework.}
The baseline 
network HAN~\cite{tian2020HAN} uses the multi-head attention (\texttt{MHA}) mechanism in Transformer~\cite{vaswani2017attention} to encode intra-modal and cross-modal features for audio and visual modalities.
We denote the initial audio and visual features extracted by pretrained convolutional neural networks as $ \bm{F}^a $, $\bm{F}^v  \in \mathbb{R}^{T \times d}$, where $d$ is the feature dimension.
The process of HAN can be summarized as,
\begin{equation}
\left\{
\begin{gathered}
\begin{split}
\bm{\dot{F}}^a = \bm{F}^a + \texttt{MHA}(\bm{F}^a, \bm{F}^a) + \texttt{MHA}(\bm{F}^a, \bm{F}^v), \\
\bm{\dot{F}}^v = \bm{F}^v + \texttt{MHA}(\bm{F}^v, \bm{F}^v) + \texttt{MHA}(\bm{F}^v, \bm{F}^a),
\end{split}
\end{gathered}
\right.
\label{eq:HAN}
\end{equation}
where $\bm{\dot{F}^a}$, $\bm{\dot{F}^v} \in \mathbb{R}^{T \times d}$ are the updated audio and visual features.
The probability of segment-wise events for audio and visual modalities $\bm{P}^a \in \mathbb{R}^{T \times C}$ and $\bm{P}^v \in \mathbb{R}^{T \times C}$ can be predicted through a fully-connected (FC) layer and a sigmoid function. 
Attentive MMIL pooling is further proposed in ~\cite{tian2020HAN} to weigh and aggregate the temporal audio-visual features. Then, we obtain the final video-level predictions of the audio events, visual events, and their union events in the entire video, denoted as $\{ \bm{p}^a; \bm{p}^v; \bm{p}^{v \cup a}\} \in \mathbb{R}^{1 \times C}$.
The objective function for model training is:
\begin{equation}\label{eq:HAN_loss}
\mathcal{L} = \mathcal{L}_{\text{bce}}(\bm{p}^{v \cup a}, \bm{y}^{v \cup a}) + \mathcal{L}_{\text{bce}}(\bm{p}^a, \overline{\bm{y}}^a) + \mathcal{L}_{\text{bce}}(\bm{p}^v, \overline{\bm{y}}^v),
\end{equation}
where $\mathcal{L}_{\text{bce}}$ is the binary cross entropy loss, $\{\overline{\bm{y}}^v, \overline{\bm{y}}^a \} \in \mathbb{R}^{1 \times C} $ are the video-level visual and audio pseudo labels generated by label smoothing~\cite{szegedy2016rethinking}  from $\bm{y}^{v \cup a}$.

\section{Our Method}
An overview of our method is shown in Fig.~\ref{fig:framework}.
We focus on producing reliable segment-level pseudo labels to better supervise the model for audio-visual video parsing.
For the backbone, we simply adopt the baseline HAN~\cite{tian2020HAN} to output predictions.
Our method provides the following new innovations. 
1) We propose a \textbf{pseudo label generation (PLG)} module that uses the pretrained CLIP model to generate reliable visual pseudo labels from the segment level.
2) To utilize the obtained pseudo label, we design a \textbf{richness-aware loss (RL)} for the model training.
This loss helps to regularize the predictions to be aware of the category richness and segment richness contained in the pseudo label.
3) We further propose a \textbf{pseudo label denoising (PLD)} strategy which can refine the generated pseudo labels by overturning the pseudo labels {of those hard-to-learn data with high forward loss values of the $\mathcal{L}_{\text{bce}}$ estimation (\ie, large binary cross-entropy)}.

\subsection{Pseudo Label Generation}\label{sec:method_PLG}
PLG aims to generate high-quality segment-level visual pseudo labels that are expected to alleviate the label interference from the video label and better supervise the model to distinguish video segments. 
As discussed in Sec.~\ref{sec:introduction}, we select the pretrained CLIP~\cite{radford2021CLIP} model to achieve this goal due to its flexible zero-shot classification capability. Here, we detail the pseudo label generation process.

Specifically, each video sample is evenly split into several video segments and we sample one frame to represent each video segment.
As shown in Fig.~\ref{fig:framework}-1, for the sampled frame $\mathcal{I}_t$ at the $t$-th segment, CLIP image encoder is used to extract the image feature, denoted as $\bm{f}^{\mathcal{I}}_t \in \mathbb{R}^{1 \times d}$.
The text input of CLIP text encoder follows the prompt ``\texttt{A photo of a [CLS]}'' where the \texttt{[CLS]} can be replaced by the potential category names. For the AVVP task, we empirically change the prompt to a more appropriate one, ``\texttt{This photo contains the [CLS]}'' (the ablation study of prompt in CLIP text encoder is shown in Sec.~\ref{exp:parameter_study}). 
By replacing the \texttt{[CLS]} in this prompt with each event category name and sending the generated texts 
to the CLIP text encoder, we can obtain the text (category) features of all $C$-class $\bm{f}^{\mathcal{T}} \in \mathbb{R}^{C \times d}$.
Then the normalized cosine similarity $\bm{s}_t \in \mathbb{R}^{1 \times C}$ between the image and event categories can be computed by,
\begin{equation}\label{eq:cosine_simmlarity}
\bm{s}_t = softmax(\frac{ \bm{f}^{\mathcal{I}}_t }{\| \bm{f}^{\mathcal{I}}_t \|_2 }  \otimes (\frac{ \bm{f}^{\mathcal{T}}}{ \| \bm{f}^{\mathcal{T}} \|_2})^{\top}),
\end{equation}
where $\otimes$ denotes the matrix multiplication, and $\top$ is the matrix transposition. A high similarity score in $\bm{s}_t$ indicates that the event category is more likely to appear in the image.

Furthermore, we use a threshold $\tau$ to select the categories with high scores in $\bm{s}_t$ and obtain the score mask $\bm{m}_t$. After that, we impose the score mask $\bm{m}_t$ on the video-level label $\bm{y}^{v \cup a}$ with element-wise multiplication $\odot$ to filter out the visual events occurring at $t$-th segment $\bm{\hat{y}}^v_t \in \mathbb{R}^{1 \times C}$. This process can be formulated as,
\begin{equation}\label{eq:PLG_tau}
\left\{
\begin{gathered}
\begin{split}
    & \bm{m}_t = \mathbbm{1}({\bm{s}_t - \tau}); \\
    & \bm{\hat{y}}^v_t = \bm{m}_t \odot \bm{y}^{v \cup a},
\end{split}
\end{gathered}
\right.
\end{equation}
where $\mathbbm{1}(x_i)$ outputs `1' when the input $x_i \geq 0$ else outputs `0', $ i=1, 2, ..., C$, and $\bm{m}_t \in \mathbb{R}^{1 \times C}$. 

This pseudo label generation process can be easily applied to all the video segments. With the proposed PLG, we obtain the segment-level pseudo label for each video, denoted as $\bm{\hat{Y}}^v =\{ \bm{\hat{y}}^v_t \}\in \mathbb{R}^{T \times C}$.  
Note that the video-level pseudo label $\hat{\bm{y}}^v \in \mathbb{R}^{1 \times C}$ can be easily obtained from $\bm{\hat{Y}}^v$, where $ \hat{{y}}^v_c =  \mathbbm{1}(\sum_{t=1}^{T} \bm{\hat{Y}}^v_{t,c})$ that means if a category of the event exists in at least one video segment, it is contained in the video-level label.

\subsection{Richness-aware Loss}

In this part, we try to regularize the model from the segment level since we have obtained the segment-level pseudo label $\hat{\bm{Y}}^v$.
This is essential for the weakly-supervised AVVP task that requires predicting for each segment, but only the video-level label is provided.
To better supervise the model, we utilize the obtained pseudo label from two aspects.

\textbf{Basic video-level loss.} Existing methods usually adopt the objective function formulated in Eq.~\ref{eq:HAN} for model training~\cite{wu2021MA,yu2022MMPyramid,cheng2022JoMOLD,mo2022MGN}, where $\overline{\bm{y}}^v \in \mathbb{R}^{1 \times C}$ is the video-level label obtained by label smoothing.
Instead, we use the video-level pseudo label $\hat{\bm{y}}^v \in \mathbb{R}^{1 \times C}$ generated by our PLG module as new supervision. The objective function is then updated to,
\begin{equation}\label{eq:RL-video_level}
\mathcal{L_V} = \mathcal{L}_{\text{bce}}(\bm{p}^{v \cup a}, \bm{y}^{v \cup a}) + \mathcal{L}_{\text{bce}}(\bm{p}^a, \overline{\bm{y}}^a) + \mathcal{L}_{\text{bce}}(\bm{p}^v, \hat{\bm{y}}^v).
\end{equation}

\textbf{Segment-level richness-aware loss.} We propose a new richness-aware loss that is inspired by the following observations.
\textbf{1)} Each row of the segment-wise pseudo label, \eg, $\hat{\bm{Y}}^v_{t\cdot} \in \mathbb{R}^{1 \times C}$,
indicates whether all the visual events appear in the $t$-th segment.
For example, as shown in Fig.~\ref{fig:framework}-2, there are three visual events in the first segment, \ie, the \textit{dog}, \textit{rooster}, and \textit{speech}, $\hat{\bm{Y}}^v_{1\cdot} = [1,1,1]$. While the last two segments only contain one \textit{dog} event, \ie, $\hat{\bm{Y}}^v_{(T-1)\cdot} = \hat{\bm{Y}}^v_{T\cdot} = [1,0,0]$.
This reflects the richness of the event category in different segments that indicates \textit{how many event categories exist in each video segment}.
We define the category richness of $t$-th segment ${cr}_t$ as the ratio of the category number of $t$-th segment to the category number of the entire video, written as, 
\begin{equation}
    {cr}_t = \frac{ \sum_{c=1}^C \hat{\bm{Y}}^v_{t,c} }{ \sum_{c=1}^C \bm{y}^{v \cup a}_{c}}
\end{equation}

Therefore, we can obtain the category richness vector of all segments $\bm{cr} \in \mathbb{R}^{T \times 1}$. 
In the example shown in Fig.~\ref{fig:framework}-2, the category richness for the first and last segments, \ie, $cr_1$ and  $cr_T$, is equal to 1 and 1/3, respectively.

\textbf{2)} Similarly, each column of the pseudo label, $\hat{\bm{Y}}^v_{\cdot c}\in \mathbb{R}^{T \times 1}$, indicates \textit{how many segments contain event of $c$-th category.}
We denote the segment richness of $c$-th category ${sr}_c$ as the ratio of the number of segments containing that category and the number of all video segments, formulated as below, 
\begin{equation}
    {sr}_c = \frac{1}{T} \sum_{t=1}^T \hat{\bm{Y}}^v_{t,c}.
\end{equation}
Extending to all $C$ event categories, we can obtain the segment richness vector of all the categories $\bm{sr} \in \mathbb{R}^{1 \times C}$.
In the example shown in Fig.~\ref{fig:framework}-2, the segment richness for the event category \textit{dog} and \textit{speech}, \ie, $sr_1$ and $sr_3$ is equal to 1 and 1/4, respectively. 
 
Given the segment-level pseudo label $\hat{\bm{Y}}^v$, we obtain the category richness $\bm{cr}$ and segment richness $\bm{sr}$.
With the visual prediction $\bm{P}^v \in \mathbb{R}^{T \times C}$ of the baseline network, we can compute the category richness and segment richness
in the same way, denoted as $\bm{pcr} \in \mathbb{R}^{T \times 1}$ and $\bm{psr} \in \mathbb{R}^{1 \times C}$, respectively.
Then we propose the richness-aware loss $\mathcal{L_S}$ to align the richness of the visual predictions and the pseudo label, calculated by,
\begin{equation}\label{eq:RL-segment_level}
   \mathcal{L_S} = \mathcal{L}_{\text{bce}}(\bm{pcr}, \bm{cr}) + \mathcal{L}_{\text{bce}}(\bm{psr}, \bm{sr}).
\end{equation}
The total objective function $\mathcal{L}_{\text{total}}$ for AVVP in this work is the combination of the basic loss $\mathcal{L_V}$ and the richness-aware loss $\mathcal{L_S}$, \ie,
\begin{equation}\label{eq:RL}
    \mathcal{L}_{\text{total}} = \mathcal{L_V} + \lambda \mathcal{L_S},
\end{equation}
where $\lambda$ is a weight parameter.

{\subsection{Pseudo Label Denoising}}\label{sec:method_PLD}



PLG can produce trustworthy pseudo labels, but it still has noises that are restricted to the pre-trained technique. For example, it is hard for PLG to identify exactly what events happen if subjects in the segment are partially occluded or poorly lit.
The quality of pseudo label is an inevitably common issue in existing label denoising methods~\cite{wu2021MA,cheng2022JoMOLD}. We thereby design a pseudo label denoising (PLD) strategy to refine the pseudo labels obtained by PLG.

Specifically, we first use the objective function shown in Eq.~\ref{eq:RL} to train a baseline model.
Then, we compute the element-wise forward propagation loss matrix $\bm{\mathcal{M}}$ between the prediction $\bm{P}^v$ and the pseudo label $\bm{\hat{Y}}^v$ for the training data: $ \bm{\mathcal{M}} = \mathcal{L}_{\text{bce}}(\bm{P}^v, \bm{\hat{Y}}^v)$, and $\bm{\mathcal{M}} \in \mathbb{R}^{T \times C}$. 
Denote the $j$-th column of $\bm{\mathcal{M}}$ as $\bm{\mathcal{M}}_{\cdot j}  \in \mathbb{R}^{T \times 1}$, it indicates the cross-entropy value of all segments for the specific $j$-th event category.
The pretrained baseline model usually has much small loss values on these segments whose pseudo labels are the same as the real ground truth.
Otherwise, segments with false pseudo labels generally have large losses and their pseudo labels should be reversed.

Note that video-level pseudo label $\bm{\hat{y}}^v \in \mathbb{R}^{1 \times C}$ 
indicates the event categories existing in the visual modality.
We use $\bm{\hat{y}}^v$ to mask the matrix $\bm{\mathcal{M}}$ to ensure label denoising 
with predicted event categories. For other event categories in the event category vocabulary that do not occur in the video, their pseudo labels will be eased by setting zeros in $\bm{\mathcal{M}}$. 
For the example shown in Fig.~\ref{fig:framework}-2, we only need to denoise the pseudo labels for the first three columns 
that corresponds to the event categories of \textit{dog}, \textit{rooster} and \textit{speech} in $\bm{\hat{y}}^v$. 
The calculation of the masked 
matrix $\bm{\mathcal{M}'}$ can be computed by, 
\begin{equation}
\begin{gathered}
\begin{split}
    & \bm{\mathcal{M}'} = f_{\text{rpt-T}}(\bm{\hat{y}}^v) \odot \bm{\mathcal{M}}
\end{split}
\end{gathered}
\end{equation}
where $\bm{\mathcal{M}'} \in \mathbb{R}^{T \times C}$, and $f_{\text{rpt-T}}(\bm{\hat{y}}^v)$ denotes the operation of repeating $\bm{\hat{y}}^v$ along the temporal dimension for $T$ times, and $f_{\text{rpt-T}}(\bm{\hat{y}}^v) \in \mathbb{R}^{T \times C}$.


For the $j$-th category, the forward loss of all the video segments is $\bm{\mathcal{M}'}_{\cdot j} \in \mathbb{R}^{T \times 1}$.
Those segments with abnormally large loss values 
may have a potentially incorrect pseudo label.
To correct the pseudo labels of such segments, we treat the average of the top-$K$ smallest loss values  of $\bm{\mathcal{M}'}_{\cdot j}$ (trustworthy prediction score) as the threshold $\mu_j$. 
By comparing the cross-entropy values of each video segment with $\mu_j$, we can obtain a binary mask vector $\bm{\varphi}_j \in \mathbb{R}^{T \times 1}$, where `1' reflects that the segment has a larger loss than $\mu_j$ and may have an incorrect pseudo label.
This process can be written as,
\begin{equation}
\left\{
\begin{gathered}
\begin{split}
    & \mu_j = f_{\text{avg}}(f_{\Bbbk}(\bm{\mathcal{M}'}_{\cdot j})) \cdot \alpha, \\
     & \bm{\varphi}_j = \mathbbm{1}(\bm{\mathcal{M}'}_{\cdot j} - \mu_j),
\end{split}
\end{gathered}
\right.
\label{eq:PLD}
\end{equation}
where $f_{\Bbbk}$ and $f_{\text{avg}}$ denotes the top-$K$ minimum loss selection and the average operation, respectively. $\alpha$ is a scaling factor to magnify the averaged loss.

Extending to all the event categories, we can obtain the $\bm{\Phi} = \{ \bm{\varphi}_j \} \in \mathbb{R}^{T \times C}$.
Finally, the pseudo label $\bm{\hat{Y}}^v$ produced by PLG can be refined by reversing the positions that have large loss values reflected by $\bm{\Phi}$, denoted as 
$ \bm{\widetilde{Y}}^v = f_{\sim}(\bm{\hat{Y}}^v, \bm{\Phi})$. 
As shown in Fig.~\ref{fig:framework}-2, for the event category \textit{rooster}, the pseudo labels generated by PLG are `0' for the last two segments that indicate the \textit{rooster} does not occur in these segments.
However, they get a large forward loss (marked by the purple box in Fig.~\ref{fig:framework}-3), compared to other segments.
This indicates the pseudo labels for the category \textit{rooster} are not correct for these two segments and are thus reversed during the denoising process.
Pseudo labels refined by PLD can be used as new supervision for the model training.

\vspace{4mm}
\section{Experiments}
\subsection{Experimental Setup}\label{exp:setup}
\textbf{Dataset.}
Experiments are conducted on the public \textit{Look, Listen, and Parse (LLP)}~\cite{tian2020HAN} dataset.
It contains 11,849 videos spanning over 25 common audio-visual categories, \eg, the scenes of humans, animals, vehicles, and musical instruments.
Each video is 10 seconds long and around 61\% of the videos contain more than one event category.
Videos of the LLP dataset are split into 10,000 for training, 649 for validation, and 1,200 for testing.
The training set is provided with only the video-level labels, \ie, the label union of the audio events and visual events.
For the validation and test sets, the segment-wise event labels for each audio and visual modality are  additionally provided.

\textbf{Evaluation metrics.}
Following existing works~\cite{tian2020HAN,cheng2022JoMOLD,wu2021MA,yu2022MMPyramid}, we evaluate the methods by measuring the parsing results of all types of events, namely audio events (\textbf{A}), visual events (\textbf{V}), and audio-visual events (\textbf{AV}, both audible and visible).
Both the segment-level and event-level F-scores are used as evaluation metrics.
The segment-level metric measures the quality of the predicted events by comparing with the ground truth for each video segment.
And the event-level metric treats consecutive segments containing the same event category as a whole event, and computes the F-score based on mIoU = 0.5 as the threshold.
The average parsing result of the three types of events is denoted as the ``\textbf{Type@AV}'' metric.
Furthermore, there is an ``\textbf{Event@AV}'' metric that considers both the predictions of the audio events and the visual events.

\textbf{Implementation details.} \textit{1) Feature extraction.} For the LLP dataset,  each 
video is divided into 10 consecutive 1-second segments.
For a fair comparison, we adopt the same feature extractors to extract the audio and visual features.
Specifically, the VGGish~\cite{hershey2017vggish} network pretrained on AudioSet dataset~\cite{gemmeke2017audioset} is used to extract the 128-dim audio features.
The pretrained ResNet152~\cite{he2016resnet} and R(2+1)D~\cite{tran2018closer} are used to extract the 2D and 3D visual features, respectively.
The low-level visual feature is the concatenation of 2D and 3D visual features. 
The ViT-based CLIP~\cite{vaswani2017attention} is used during the pseudo label generation, and the parameters are frozen. 
\textit{2) Training procedure}.
For each video in the training set of the LLP dataset, we first offline generate the segment-wise visual pseudo labels using our PLG module.
Then, the objective function $\mathcal{L}_{\text{total}}$ shown in Eq.~\ref{eq:RL} is used to train the baseline model HAN~\cite{tian2020HAN}.
The hyperparameter $\lambda$ in Eq.~\ref{eq:RL} for balancing the video-level and the segment-level losses is empirically set to 0.5. 
This pretrained model is then used in our PLD to further refine the pseudo labels.
Finally, the pseudo labels obtained by PLD are used to supervise the baseline model training again.
For all the training processes, we adopt the Adam optimizer to train the model with a mini-batch size of 32 and the learning rate of $3 \times 10^{-4}$. The total training epoch is set to 30.
All experiments are conducted with PyTorch~\cite{paszke2019pytorch} on one NVIDIA GeForce-RTX-2080-Ti GPU.
Codes will be released.

\begin{table}[t]\small
\caption{\textbf{Parameter study of $\tau$ and prompt used in the PLG.}
Different setups are used to generate segment-level visual pseudo labels, making it applicable to also obtain visual pseudo labels at the video level. In this table, we report the precision, recall, and average result (\%) of visual track between the pseudo label and the ground truth at the video level.
\textit{`-' denotes the result of directly assigning video labels as the visual event labels.}
The segment-level performance will be discussed in Tabel~\ref{tab:parameter_study_PLD}. 
This experiment is conducted on the validation set of the LLP dataset.} 
\centering
\label{tab:parameter_study_PLG}
\small
 \begin{tabular}{p{1.3cm}<{\centering}p{1.3cm}<{\centering}p{1.2cm}<{\centering}p{1.2cm}<{\centering}p{1.2cm}<{\centering}}
\toprule[0.8pt]
\multicolumn{2}{c}{Parameter setup} & \multicolumn{1}{c}{\multirow{2}{*}{Precision (V)}}  & \multicolumn{1}{c}{\multirow{2}{*}{Recall (V)}}  & \multicolumn{1}{c}{\multirow{2}{*}{Average}}  \\ \cmidrule{1-2}
\multicolumn{1}{c}{$\tau$} & \multicolumn{1}{c}{prompt} & & & \\ \midrule
- & - & 66.96 & 87.88 & 77.42 \\
\midrule
0.035 & \multirow{4}{*}{P1} & 83.69 & 81.15 & 82.42 \\
\textbf{0.040} & & 83.77 & 81.15 & \textbf{82.46} \\
{0.041} &  & {81.33} & 77.38 & {79.35} \\
0.043 &  & 38.29 & {34.14} & 36.22 \\ \midrule
\multirow{4}{*}{0.040} & \textbf{P1} & 83.77 & 81.15 & \textbf{82.46} \\
 & P2 & 80.12 & 75.46 & 77.79 \\
 & P3 & 78.22 & {72.74} & 75.48 \\
 & P4 & {83.08} & 79.66 & 81.37 \\
\toprule[0.8pt]
\end{tabular}
\end{table}


\subsection{Parameter Studies}\label{exp:parameter_study}
We first perform parameter studies of essential parameters used in our method, \ie, the mask threshold $\tau$ and CLIP prompt used in the PLG module, the top-$K$ and scaling factor $\alpha$ used in the PLD strategy.
Experiments in this section are conducted on the validation set of the LLP dataset of which the segment-level event label is known for parameter tuning. Thus, we also discuss \textbf{the measurement of pseudo label correctness} in this part.

\textbf{Study of $\tau$ and prompt in PLG.}
$\tau$ is used as a threshold to select high scores of the cosine similarity between the event category and the video segment in the mask calculation (Eq.~\ref{eq:PLG_tau}).
To explore its impact, we fix the prompt of CLIP to \textbf{P1} -- ``\texttt{This photo contains the [CLS]}'' and test several values of $\tau$ to produce the segment-level visual pseudo labels, 
making it applicable to obtain visual pseudo labels at the video level too. Then, \textit{we calculate the \textit{Precision} and \textit{Recall} between the pseudo labels and the ground truth on the validation set at video-level} and use the average metric to select the best setup. As shown in the upper part of Table~\ref{tab:parameter_study_PLG}, the best average result is achieved when $\tau = $ 0.04. The performance is relatively stable while $\tau \in [0.035, 0.04]$, 
and the precision drops significantly when $\tau$ changes from 0.041 to 0.043. We argue such sensitivity is related to the \emph{softmax} operation in Eq.~\ref{eq:cosine_simmlarity} that squeezes the similarity score into small logits.

Furthermore, we explore the impact of prompt in the CLIP.
Specifically, we test four types of prompts, \ie, our default \textbf{P1} -- ``\texttt{This photo contains the [CLS]}'', \textbf{P2} -- ``\texttt{This photo contains the scene of [CLS]}'', \textbf{P3} -- ``\texttt{This photo contains the visual scene of [CLS]}'' and \textbf{P4} -- ``\texttt{This is a photo of the [CLS]}''.
We use these different prompts to generate the pseudo labels and compare them with the ground truth. 
As shown in the lower part of Table~\ref{tab:parameter_study_PLG}, pseudo labels generated using these different prompts keep relatively consistent.
The pseudo label has the highest average score using the \textbf{P1} prompt.
Therefore, we set $\tau$ to 0.04 and use the prompt \textbf{P1} as the optimal setup in our following experiments.
Notably, the precision of the video-level pseudo label reaches about 84\% under the optimal setup.
Whereas the precision of directly assigning video labels as the visual event labels is only $\sim$67\%. 
This reveals that PLG can satisfactorily disentangle visual events from weak video labels.
The segment-level evaluation of the pseudo labels will be discussed in Tabel~\ref{tab:parameter_study_PLD}.

\begin{table}[t]\small
\caption{\textbf{Parameter study of the $K$ and $\alpha$ used in the PLD.} 
Different values of $K$ and $\alpha$ are tested for the segment-wise pseudo label denoising.
The segment-level and event-level F-scores of the denoised visual pseudo labels are reported. Notably, in this table, we evaluate the pseudo labels of the visual track (V). The last column is the average result. \textit{`-' denotes the result of the pseudo label generated by PLG without label denoising.} This experiment is conducted on the validation set of the LLP dataset.
} 
\centering
\label{tab:parameter_study_PLD}
\small
 \begin{tabular}{p{1.2cm}<{\centering}p{1.2cm}<{\centering}p{1.2cm}<{\centering}p{1.2cm}<{\centering}p{1.2cm}<{\centering}}
\toprule[0.8pt]
\multicolumn{2}{c}{Parameter setup} & \multicolumn{1}{c}{\multirow{2}{*}{Segment. (V)}}  & \multicolumn{1}{c}{\multirow{2}{*}{
Event. (V)}}  & \multicolumn{1}{c}{\multirow{2}{*}{Average}}  \\ \cmidrule{1-2}
\multicolumn{1}{c}{$K$} & \multicolumn{1}{c}{$\alpha$} & & & \\ \midrule
- & - & 70.29 & 64.68 & 67.49 \\ \midrule
2 & \multirow{4}{*}{10} & 72.45 & 67.13 & 69.79 \\
3 & & 72.26 & 67.30 & 69.78 \\
\textbf{5} & & \textbf{72.51} & \textbf{68.09} & \textbf{70.30} \\
6 & & 71.84 & 66.91 & 69.38 \\ \midrule
\multirow{3}{*}{5} & 5 & 72.09 & 66.44 & 69.26 \\
& \textbf{10} & \textbf{72.51} & \textbf{68.09} & \textbf{70.30} \\
& 20 & 72.05 & 67.58 & 69.81 \\
\toprule[0.8pt]
\end{tabular}
\end{table}

\textbf{Study of the $K$ and $\alpha$ in PLD.}
In the pseudo label denoising, 
for each predicted event category, the top-$K$ smallest forward loss along the temporal dimension magnified by $\alpha$ is used as the threshold to determine which segments' pseudo labels should be corrected.
To explore its impact, we test various combinations of these two parameters. 
\textit{The segment-level and event-level F-scores of the visual events are used to measure the quality of pseudo label after denoising.}
The results in Table~\ref{tab:parameter_study_PLD} indicate that pseudo labels after denoising ensure better performance results than the original label generated by PLG.
The optimal setup is $K = $ 5, and $\alpha = $ 10.
For the visual pseudo labels of the validation set, the segment-level and event-level F-scores are 72.51\% and 68.09\%, respectively, under the optimal setup.
We cannot exactly compute these F-scores for the training set as it does not provide segment-level labels.
However, we speculate that the F-scores for the training set should be not bad or even better than the validation set (segment-level and event-level F-scores 72.51\% and 68.09\%) since these two sets share similar data distribution from one dataset and we use the same PLG module to obtain the pseudo labels.
Notably, observing to the test set, the current state-of-the-art visual event parsing performance is 58.3\% F-score at the segment level, achieved by method DHHN~\cite{jiang2022DHHN}.
This gap reveals that the pseudo labels of the training set generated by PLG and refined by PLD can serve as a reliable target for model learning.

\begin{table*}[t]\small
\caption{\textbf{Ablation study of the main modules.}
$\mathcal{L_S}$ is the proposed richness-aware loss (Eq.~\ref{eq:RL-segment_level}) and 
$\mathcal{L'_S}$ is a native loss that simply computes the binary cross entropy loss of the prediction and pseudo label.
We report the results on the test set of the LLP dataset.}
\centering
\label{tab:ablation}
 \begin{tabular}{p{0.2cm}<{\centering}p{0.5cm}<{\centering}p{0.8cm}<{\centering}p{0.5cm}<{\centering}p{0.9cm}<{\centering}p{0.9cm}<{\centering}p{0.9cm}<{\centering}p{1.2cm} <{\centering}p{1.2cm} <{\centering}p{0.9cm}<{\centering}p{0.9cm}<{\centering}p{0.9cm}<{\centering}p{1.2cm}<{\centering}p{1.2cm}<{\centering}}
\toprule[0.8pt]
 \multirow{2}{*}{Id} & \multicolumn{3}{c}{Main modules} & \multicolumn{5}{c}{Segment-level} & \multicolumn{5}{c}{Event-level}  \\
 \cmidrule(r){2-4} \cmidrule(r){5-9} \cmidrule{10-14}
  & PLG & RL & PLD & A & V & AV & Type@AV & Event@AV & A & V & AV & Type@AV & Event@AV \\
\midrule
\ding{192} & \ding{56} & \ding{56} &   \ding{56} & 60.1 & 52.9 & 48.9 & 54.0 & 55.4 & 51.3 & 48.9 & 43.0 & 47.7 & 48.0 \\
\midrule
 \ding{193} & \ding{52} & \ding{56} &   \ding{56} & 60.0 & 63.7 & 57.3 & 60.3 & 58.8 & 51.1 & 60.3 & 50.4 & 53.9 & 50.2 \\
\midrule
 \ding{194} & \ding{52}  & \ding{52}-$\mathcal{L'_S}$ & \ding{56} & 60.2 & {64.1} & 57.6 & 60.6 & 59.0 & 50.8 & 60.5 & 50.1 & 53.8 & 49.8 \\
\ding{195} & \ding{52}  & \ding{52}-$\mathcal{L_S}$ & \ding{56} & 60.5 & 64.8 & 58.3 & 61.2 & 59.4 & 51.4 & 61.5 & 51.2 & 54.7 & 50.8 \\
 \midrule
\ding{196} &  \ding{52}  & \ding{52} & \ding{52} & \textbf{61.1} & \textbf{65.5} & \textbf{59.3} & \textbf{62.0} & \textbf{60.1} & \textbf{52.1} & \textbf{62.4} & \textbf{52.4} & \textbf{55.6} & \textbf{51.3}  \\ 
\toprule[0.8pt]
\end{tabular}
\end{table*}

\begin{table*}[t]\small
\caption{\textbf{Comparison with the state-of-the-arts.}
$^{\diamondsuit}$ represents these methods are all focused on generating better pseudo labels for the AVVP task and are all developed on the baseline HAN~\cite{tian2020HAN} framework.
Results are reported on the test set of the LLP dataset.}
\centering
\label{tab:sota_comparison}
 \begin{tabular}{p{3.4cm}<{\centering} p{0.8cm}<{\centering}p{0.8cm}<{\centering}p{0.8cm}<{\centering}p{1.2cm} <{\centering}p{1.2cm} <{\centering}p{1.cm}<{\centering}p{0.8cm}<{\centering}p{0.8cm}<{\centering}p{1.2cm}<{\centering}p{1.2cm}<{\centering}}
\toprule[0.8pt]
\multirow{2}{*}{Method} & \multicolumn{5}{c}{Segment-level} & \multicolumn{5}{c}{Event-level}  \\
 \cmidrule(r){2-6} \cmidrule(r){7-11}
 & A & V & AV & Type@AV & Event@AV & A & V & AV & Type@AV & Event@AV \\
\midrule
AVE~\cite{tian2018audio} & 47.2 & 37.1 & 35.4 & 39.9 & 41.6 & 40.4 & 34.7 & 31.6 & 35.5 & 36.5 \\
AVSDN~\cite{lin2019dual} & 47.8 & 52.0 & 37.1 & 45.7 & 50.8 & 34.1 & 46.3 & 26.5 & 35.6 & 37.7 \\
\rowcolor{gray!20}HAN~\cite{tian2020HAN} & 60.1 & 52.9 & 48.9 & 54.0 & 55.4 & 51.3 & 48.9 & 43.0 & 47.7 & 48.0 \\
MM-Pyramid~\cite{yu2022MMPyramid} & 60.9 & 54.4 & 50.0 & 55.1 & 57.6 & 52.7 & 51.8 & 44.4 & 49.9 & 50.5 \\
MGN~\cite{mo2022MGN} & 60.8 & 55.4 & 50.4 & 55.5 & 57.2 & 51.1 & 52.4 & 44.4 & 49.3 & 49.1 \\
CVCMS~\cite{lin2021CVCMS} & 59.2 & 59.9 & 53.4 & 57.5 & 58.1 & 51.3 & 55.5 & 46.2 & 51.0 & 49.7 \\
DHHN~\cite{jiang2022DHHN} & 61.3 & 58.3 & 52.9 &  57.5 & 58.1 & 54.0 & 55.1 & 47.3 & 51.5 & 51.5 \\
\midrule
$^{\diamondsuit}${MA}~\cite{wu2021MA} & 60.3 & 60.0 & 55.1 & 58.9 & 57.9 & 53.6 & 56.4 & 49.0 & 53.0 & 50.6 \\
$^{\diamondsuit}$JoMoLD~\cite{cheng2022JoMOLD} & \textbf{61.3} & 63.8 & 57.2 & 60.8 & 59.9 & \textbf{53.9} & 59.9 & 49.6 & 54.5 & \textbf{52.5} \\ 
$^{\diamondsuit}$\textbf{VPLAN (ours)} & 61.1 & \textbf{65.5} & \textbf{59.3} & \textbf{62.0} & \textbf{60.1} & 52.1 & \textbf{62.4} & \textbf{52.4} & \textbf{55.6} & 51.3 \\ 
\toprule[0.8pt]
\end{tabular}
\end{table*}

\subsection{Ablation Studies}\label{exp:ablations}
In this section, we provide some ablation studies to explore the impact of each module in our method.
The experimental results are shown in Table~\ref{tab:ablation}.
The row with id-\ding{192} denotes the performance of the baseline HAN~\cite{tian2020HAN}.

\textbf{Impact of the PLG.}
With the PLG module, we obtain the pseudo labels from both segment-level and video-level, denoted as $\hat{\bm{Y}}^v$ and $\hat{\bm{y}}^v$, respectively.
It is noteworthy that id-\ding{192} (the basic HAN) only uses the video-level pseudo label $\overline{\bm{y}}^v$ 
for model training (Eq.~\ref{eq:HAN_loss}). For a fair comparison, we use the video-level pseudo label $\hat{\bm{y}}^v$ to supervise the model training (Eq.~\ref{eq:RL-video_level}). 
As shown in row-\ding{193} of Table~\ref{tab:ablation}, utilizing the pseudo label generated by PLG significantly improves the performance of visual-related event prediction. The average metric Type@AV increases from 54.0\% to 60.3\% at the segment-level while from 47.7\% to 53.9\% at the event-level. These results demonstrate the effective design of the proposed PLG that successfully borrows free but reliable prior knowledge (pseudo labels) from the big pretrained model.


\textbf{Impact of the Richness-aware Loss (RL).}
The loss $\mathcal{L_S}$ in Eq.~\ref{eq:RL-segment_level} is proposed to utilize the segment-level pseudo label, as a complement to the video-level supervision. To validate its effectiveness, we compare it
with a native variant that directly computes the binary cross entropy loss between the prediction and the pseudo label, denoted as $\mathcal{L'_S} = \mathcal{L}_{\text{bce}}(\bm{P}^v, \hat{\bm{Y}}^v)$.
As shown in the row-\ding{194} and \ding{195} of Table~\ref{tab:ablation}, the proposed RL $\mathcal{L_S}$ is more helpful than $\mathcal{L'_S}$ for video event parsing. For example, the metrics Type@AV in both segment-level and event-level are improved by around 1\%.
The reason may be that $\mathcal{L'_S}$ forcefully aligns the prediction and the pseudo label for each segment. Instead, $\mathcal{L_S}$ provides a soft supervision that optimizes the model by being aware of the richness information. 


\begin{figure*}[t]
\centering
\includegraphics[width=0.82\textwidth]{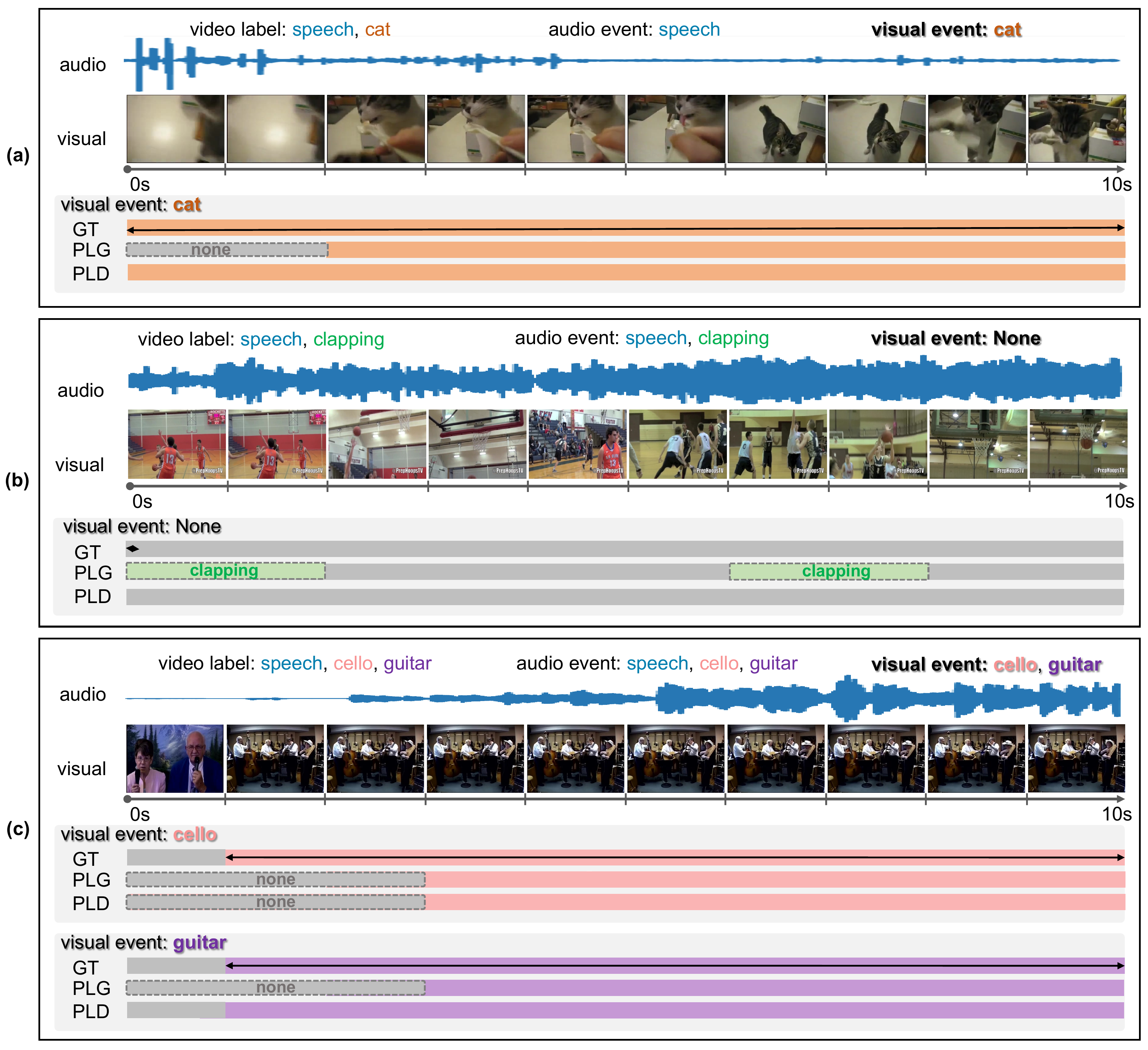}
\caption{\textbf{Visualization of the pseudo labels.}
``GT'' is the ground truth.
``PLG'' and ``PLD'' denotes the pseudo labels obtained by PLG and PLD, respectively.
PLG basically disentangles the visual event(s) from the weak video label and provides segment-wise event categories, while L-PLD further corrects the pseudo labels. 
}
\label{fig:vis_pseudo_labels}
\end{figure*}

\begin{table*}[!htp]\small
\caption{\textbf{Generalization of our method.}
Our method can generate reliable pseudo labels which can be used for other methods in the AVVP task.
We evaluate two SOTA networks, \ie, MM-Pyramid~\cite{yu2022MMPyramid} and MGN~\cite{mo2022MGN}.
The pseudo labels generated by our PLG and refined by our PLD consistently boost these models and are also superior to the existing method MA~\cite{wu2021MA}.
$^*$ denotes the results are reproduced by us using the official source code.}
\centering
\label{tab:PLG_PLD_on_other_methods}
 \begin{tabular}{p{3.6cm}<{\centering} p{0.8cm}<{\centering}p{0.8cm}<{\centering}p{0.8cm}<{\centering}p{1.2cm} <{\centering}p{1.2cm} <{\centering}p{1.cm}<{\centering}p{0.8cm}<{\centering}p{0.8cm}<{\centering}p{1.2cm}<{\centering}p{1.2cm}<{\centering}}
\toprule[0.8pt]
\multirow{2}{*}{Method} & \multicolumn{5}{c}{Segment-level} & \multicolumn{5}{c}{Event-level}  \\
 \cmidrule(r){2-6} \cmidrule(r){7-11}
 & A & V & AV & Type@AV & Event@AV & A & V & AV & Type@AV & Event@AV \\
\midrule
$^*$MM-Pyramid~\cite{yu2022MMPyramid} & {60.5} & {52.3} & 48.1 & 53.6 & 57.5 & {51.5}& 48.3 & 42.5 & 47.4 & 48.5 \\
$^*$MM-Pyramid + MA~\cite{wu2021MA} & 58.6 & \textbf{57.3} & 52.1 & 56.0 & 57.2 & 50.5 & \textbf{53.1} & 45.9 & 49.9 & {48.5} \\
MM-Pyramid + \textbf{PLG} & 59.8 & \textbf{63.9} & {57.3} & {60.3} & {59.1} & 50.0 & \textbf{60.3} & {50.2} & {53.5} & 48.6 \\
MM-Pyramid + \textbf{PLD} & 60.1 & \textbf{64.8} & 58.0 & 61.0 & 59.5 & 51.4 & \textbf{60.7} & 50.7 & {54.2} & 49.7 \\
\midrule
MGN~\cite{mo2022MGN} & {60.8} & 55.4 & 50.4 & 55.5 & 57.2 & {51.1 }& 52.4 & 44.4 & 49.3 & 49.1 \\
MGN + MA~\cite{wu2021MA} & 60.2 & \textbf{61.9} & 55.5 & 59.2 & 58.7 & 50.9 & \textbf{59.7} & 49.6 & 53.4 & {49.9} \\
MGN + \textbf{PLG} & 60.3 & \textbf{63.0} & {56.4} & {59.9} & {59.1} & 50.5 & \textbf{61.1} & {50.4} & {54.0} & 49.5 \\
MGN + \textbf{PLD} & 59.6 & \textbf{64.0} & 56.9 & 60.2 & 58.9 & 49.8 & \textbf{61.8} & 50.5 & {54.0} & 49.3 \\
\toprule[0.8pt]
\end{tabular}
\end{table*}

\textbf{Impact of the PLD.}
The impact of PLD can be observed from two aspects.
On one hand, PLD can provide more accurate pseudo labels than PLG. As shown in Table~\ref{tab:parameter_study_PLD} on the validation set, the average F-score is 67.49\% for PLG while 70.30\% for PLD. 
On the other hand, PLD is more helpful than PLG for model training. We replace PLG with PLD as the new supervision to train the HAN model.
As shown in row-\ding{196} of Table~\ref{tab:ablation}, 
the model has superior performance on most of the metrics using PLD. This reveals that the pseudo labels obtained by PLD are more accurate than by PLG and thus provide much better supervision. Also, this verifies the effectiveness of the label denoising strategy in PLD.

\begin{figure*}[!htp]
\centering
\includegraphics[width=0.8\textwidth]{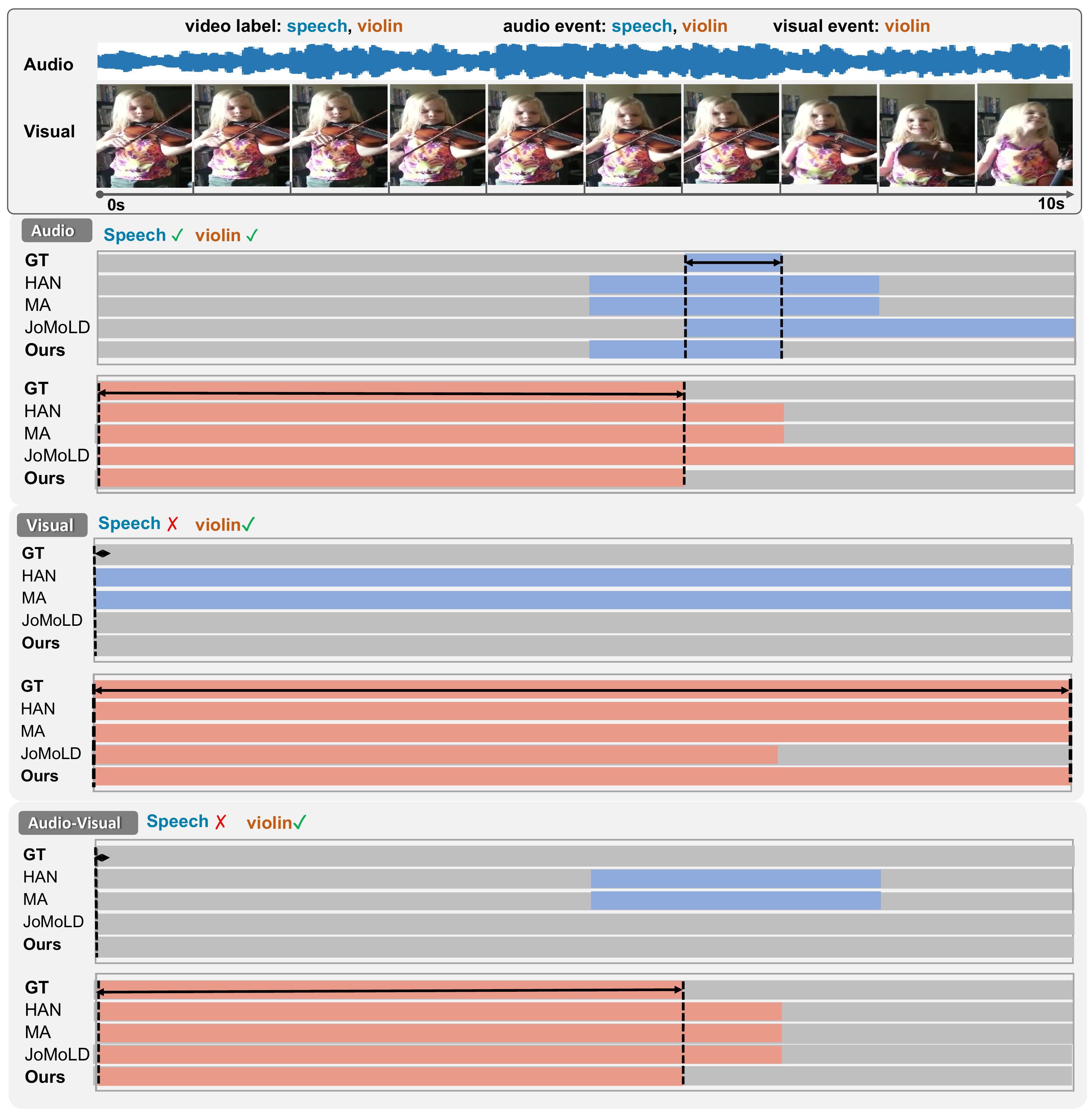}
\vspace{2mm}
\caption{\textbf{Qualitative example of the audio-visual video parsing.}
We compare our method with the HAN~\cite{tian2020HAN}, MA~\cite{wu2021MA} and JoMoLD~\cite{cheng2022JoMOLD}.
``GT'' denotes the ground truth.
Our method successfully recognizes that there is only one event \textit{violin} in the visual track and is also more accurate in parsing the audio events and audio-visual events.
}
\label{fig:vis_parsing_results}
\end{figure*}

\subsection{Comparison with the State-of-the-arts}
We perform our VPLAN method with the optimal parameters studied in Sec.~\ref{exp:parameter_study} and report the performance on the test set of the LLP dataset.
The comparison result with existing methods is shown in Table~\ref{tab:sota_comparison}.
Our method achieves overall superior performance.
\textbf{First}, compared to the baseline HAN~\cite{tian2020HAN}, our method significantly improves the performance, especially on the visual-related metrics.
Take the visual event prediction (\textbf{V} in the table) as an example, the segment-level metric is lifted from 52.9\% to 65.5\% ($\uparrow$ 12.6\%) and the event-level metric is improved from 48.9\% to 62.4\% ($\uparrow$ 13.5\%).
\textbf{Second}, our method also outperforms others on the track of generating pseudo labels for the AVVP task.
As shown in the low part of the Table, our method has comparable performance in audio event parsing but generally exceeds the method MA~\cite{wu2021MA} by $\sim$4 points and method JoMoLD~\cite{cheng2022JoMOLD} by $\sim$2 points for the visual event and audio-visual event parsing. These results demonstrate the superiority of our whole method design.

\subsection{Generalization of Our Method}
A core contribution of our method is that it can provide high-quality pseudo labels, which then better guide the model optimization.
The method can also be directly applied to existing methods in the AVVP task. To explore its impact, we examine two recently proposed methods, \ie, MM-Pyramid~\cite{yu2022MMPyramid} and MGN~\cite{mo2022MGN}.
Specifically, we train the models using the pseudo labels generated by our PLG and refined by our PLD, respectively. The experimental results are shown in Table~\ref{tab:PLG_PLD_on_other_methods}.
Both PLG and PLD significantly boost the vanilla models, especially in the visual event and audio-visual event parsing. Take the MM-Pyramid~\cite{yu2022MMPyramid} method for example, the segment-level visual event parsing performance is improved from 52.3\% to 63.9\% and 64.8\% by using PLG and PLD, respectively.
PLD is superior contributed to the proposed label denoising strategy.
Such improvements can also be observed for MGN~\cite{mo2022MGN}.
It is worth noting that these two models also perform better when combined with our pseudo labels than those generated by MA~\cite{wu2021MA}.
These results further indicate the superiority and the generalizability of our method.



\subsection{Qualitative Results}\label{exp:qualitative}
\textbf{Visualization of the pseudo labels and label denoising.} We first provide three visualization examples of the pseudo labels generated by the PLG and PLD.
As shown in Fig.~\ref{fig:vis_pseudo_labels} (a), this video contains the events of \textit{speech} and \textit{cat}, and \textit{speech} is an interference for the visual modality. 
PLG successfully recognizes that only \textit{cat} event happens in the visual track and produces the segment-wise pseudo labels. Most pseudo labels generated by PLG are correct. However, since  the object is too blurry in the first two segments, the event \textit{cat} is incorrectly recognized. Contributed to the proposed label denoising strategy, PLD makes the correction.
In Fig.~\ref{fig:vis_pseudo_labels} (b), there are no visual events.
PLG mistakenly classifies a few segments as the event \textit{clapping} because the player's movements are complex in these segments. PLD corrects the wrong predictions too.
In Fig.~\ref{fig:vis_pseudo_labels} (c), the visual track contains two events, \ie, \textit{cello} and \textit{guitar}.
Both PLG and PLD fail to recognize the \textit{cello} event in the second and third segments but PLD successfully revises the \textit{guitar} in these segments.
Generally, pseudo labels generated by PLG rely on the prior knowledge from the 
the pretrained CLIP, while PLD benefits from an additional 
revision process (-- the joint exploration of the baseline HAN~\cite{tian2020HAN} training and the forward loss $\mathcal{L}_{\text{bce}}$ calculation can correct possible false pseudo labels in PLG).

\textbf{Visualization example of the audio-visual video parsing.} We also display a qualitative video parsing example in Fig.~\ref{fig:vis_parsing_results}.
We compare our method with HAN~\cite{tian2020HAN}, MA~\cite{wu2021MA}, and JoMoLD~\cite{cheng2022JoMOLD}.
Both MA and JoMoLD are developed on the HAN and try to generate video-level pseudo labels for better model training.
As shown in the figure, two events exist in the video, \ie, \textit{speech} and \textit{violin}, while the visual event only contains the \textit{violin}.
For the audio event parsing, all the methods recognize that both two events occur in the audio track, but our method better locates more exact temporal segments.
For the visual event parsing, HAN and MA 
incorrectly predict that both events \textit{speech} and \textit{violin}  occur in the visual track. 
Although JoMoLD successfully identifies that there is only one visual event \textit{violin}, it incorrectly predicts the last three video segments.
In contrast, our method can exactly parse the visual event. 
Our method also performs better for the final audio-visual event parsing. This superiority comes from the fact that our method can generate high-quality segment-level pseudo labels, providing better supervision for model training. 


\section{Conclusion}
We propose a Visual Pseudo LAbel exploratioN (VPLAN) method for the weakly supervised audio-visual video parsing task. 
VPLAN is the first attempt to generate segment-level pseudo labels in this field, which starts with a pseudo label generation module that uses the strong CLIP to determine the events occurring in the visual modality (at the segment level) as pseudo labels.
In this work, we study the category richness and segment richness contained in the pseudo labels and propose a new richness-aware loss to constrain the label alignment.
Finally, we propose a pseudo label denoising strategy to refine the pseudo labels and better guide the predictions. 
Qualitative and quantitative experimental results on the LLP dataset corroborate that our method can generate and effectively exploit high-quality pseudo labels. 
All these proposed techniques can be directly used for the community.


\section*{Acknowledgments}
We would like to thank Dr. Liang Zheng and Dr. Yunfeng Diao for their constructive comments and suggestions.
This work was supported by the National Natural Science Foundation of China (72188101, 61725203, 62020106007, and 62272144), and the Major Project of Anhui Province (202203a05020011).

\bibliographystyle{IEEEtran}
\bibliography{IEEEabrv,IEEEtran}

\end{document}